\documentclass[11pt,a4paper]{article}
\usepackage[hyperref]{acl2021}
\usepackage{times}
\usepackage{latexsym}
\usepackage{amsmath}
\usepackage{numprint}
\usepackage{pgfplots}

\pgfplotsset{compat=1.8}
\usepgfplotslibrary{statistics}

\usepackage{amssymb}
\usepackage{booktabs}
\usepackage{graphicx}
\usepackage{subcaption}

\usepackage{microtype}

\aclfinalcopy 

\newcommand{\Tref}[1]{Table~\ref{#1}}

\newcommand{\gmu}{$^1$}
\newcommand{\naver}{$^2$}
\newcommand{\fb}{$^3$}
\newcommand{\jhu}{$^4$}

\usepackage{xcolor}

\title{On the Evaluation of Machine Translation for Terminology Consistency}

\author{Md Mahfuz ibn Alam\gmu~\;~Antonios Anastasopoulos\gmu~\;~Laurent Besacier\naver~\;~James Cross\fb \\
\textbf{Matthias Gall\'e}\naver~\;~\textbf{Philipp Koehn}\fb$^,$\jhu~\;~\textbf{Vassilina Nikoulina}\naver \\
\gmu~Department of Computer Science, George Mason University\\
\naver~NAVER Labs, Grenoble\\
\fb~Facebook, 
\jhu~Johns Hopkins University\\
\texttt{\{malam21,antonis\}@gmu.edu,   } \\
\texttt{\{laurent.besacier,matthias.galle,vassilina.nikoulina\}@naverlabs.com}
}

\date{}

\begin{document}
\maketitle
\begin{abstract}
As neural machine translation (NMT) systems become an important part of professional translator pipelines, a growing body of work focuses on combining NMT with terminologies. In many scenarios and particularly in cases of domain adaptation, one expects the MT output to adhere to the constraints provided by a terminology. In this work, we propose metrics to measure the consistency of MT output with regards to a domain terminology. We perform studies on the COVID-19 domain over 5 languages, also performing terminology-targeted human evaluation. We open-source the code for computing all proposed metrics.\footnote{\url{https://github.com/mahfuzibnalam/terminology_evaluation}}
\end{abstract}

\section{Introduction}

Commercial MT tools allow translators to leverage previously translated data, as well as hand-curated translation memories and terminologies. In fact, quite often, a client will provide a LSP (language service provider) with a domain-specific terminology to ensure that the translations produced are \textit{consistent} and follow domain-specific conventions.

In the case of newly emerging domains, in addition, terminologies are the easiest type of parallel data that one could collect. For instance, when the TICO-19 dataset~\cite{anastasopoulos-etal-2020-tico} with data on the COVID-19 domain was created, within 2 weeks Google and Facebook were able to quickly create terminologies in 100+ languages with more than 300+ translated terms each. For comparison, it took 7-8 months to translate the around 3,000 sentences that comprise the dataset in just 35 languages.
Thus, the creation (and use) of domain-specific terminologies could be a viable path towards the rapid adaptation of MT systems in case of emergencies.

Phrase-based statistical MT systems~\cite{koehn2003statistical} allowed for fine-grained control over the system's output by design, e.g. by incorporating domain-specific dictionaries into the phrase table, or by forcing translation choices for certain words or phrases. On the other hand, the currently state-of-the-art approach of neural machine translation (NMT) does not inherently allow for such control over the system's output. 
Some approaches incorporate dictionaries through interpolation of the decoder's probability with a lexical probability based on source-side attention matches~\cite{arthur-etal-2016-incorporating}. Perhaps the most common paradigm is \textit{constrained decoding}~\cite{hokamp-liu-2017-lexically,anderson-etal-2017-guided,post-vilar-2018-fast}, where the terminology matches are presented as \textit{hard} constraints that the beam search must satisfy.

Constrained decoding is not without disadvantages: it can be computationally expensive and it is often brittle when applied in realistic conditions~\cite{dinu-etal-2019-training}. To this end, \citet{dinu-etal-2019-training} and \citet{bergmanis2021facilitating} introduced an approach where the terminological constraints are provided as input to the NMT as additional annotations inline with the source sentence. As such, these can be considered as ``soft" constraints, as there is no guarantee that the NMT system will indeed produce an output containing them.

Regardless of the choice of the method for imposing constraints or incorporating terminologies in the MT model, most of the previous scholarly work on the field fails to actually evaluate whether the MT models are consistent in translating terms. All previous works evaluate the MT outputs using standard corpus-level or sentence-level metrics like BLEU~\cite{papineni2002bleu}.\footnote{\citet{bergmanis2021facilitating} do use terminology-targeted human evaluation, which albeit being a good solution, is hard and expensive to do at scale.} While such metrics can measure the general quality of the MT output, they do not explicitly measure whether the constraints are used and/or whether they are used \textit{correctly} in the MT output.

This work aims to fill this gap, by proposing metrics that will evaluate the consistency of MT output given a terminology. Beyond introducing such metrics, we apply them over several MT systems (constrained and not) and language pairs, focusing on the new, critical domain of COVID-19 data.

\section{Evaluating the consistency of MT given a terminology}

In this setup, we assume a parallel terminology and a test set of parallel sentences which contain terms that also appear in the terminology. The terms can go beyond single words to contain multi-word expressions. We show an example English-Spanish terminology in \Tref{tab:terminology}, provided as part of the TICO-19 dataset. 

\begin{table}[t]
    \centering
    \begin{tabular}{@{}p{3.5cm}|p{3.6cm}@{}}
    \toprule
    \textbf{English} & \textbf{Spanish} \\
    \midrule
1918 flu & La gripe de 1918 \\
acute bronchitis&Bronquitis aguda \\
acute respiratory disease&enfermedades respiratorias agudas \\
AIDS&SIDA \\
airborne droplets&gotas suspendidas en el aire \\
airway&vía respiratoria \\
alcohol-based&a base de alcohol \\
alcohol-based hand rub&desinfectante de manos a base de alcohol \\
\ldots & \ldots \\
\bottomrule
\end{tabular}
    \caption{Example parallel terms from the TICO-19 (related to COVID-19) English-Spanish terminology.}
    \label{tab:terminology}
\end{table}

\begin{table*}[t]
    \centering
    \begin{tabular}{@{}l|p{12cm}@{}}
    \toprule
        English & Comparably , a demographic study in 2012 showed that MERS-CoV patients also had \textbf{[fever]$_1$} (98\%) , \textbf{[dry [cough]$_2$]$_3$} (47\%) , and dyspnea (55\%) as their main \textbf{[symptoms]$_4$} .  \\
        Spanish & De manera comparable , un estudio demográfico de 2012 demostró que los pacientes con MERS-CoV tambi\'en tenían \textbf{[fiebre]$_1$} (98\%) , \textbf{[[tos]$_2$ seca]$_3$} (47\%) y disnea (55\%) como principales \textbf{[síntomas]$_4$} . \\
        Term. Entries & fever--fiebre, cough--tos, dry cough--tos seca, symptom--s\'intoma \\
        \midrule
        MT Output 1 & Comparablemente , un estudio demográfico realizado en 2012 mostró que los pacientes con MERS-CoV tambi\'en ten\'ian \textbf{fiebre} (98\%) , \textbf{tos seca} (47\%) y disnea (55\%) como sus principales \textbf{s\'intomas} .\\
        \midrule
        MT Output 2 & Comparablemente , un estudio demográfico realizado en 2012 mostró que los pacientes con MERS-CoV tambi\'en ten\'ian \textbf{fiebre} (98\%) , \textbf{tos} (47\%) y disnea (55\%) como sus principales \textbf{s\'intomas} .\\    
        \bottomrule
    \end{tabular}
    \caption{Example term annotation on the dataset for both source and target (also showing corresponding MT outputs).}
    \label{tab:example}
\end{table*}

\subsection{Data Requirements}

We will assume that we have a terminology $\mathcal{T}$ with source-target term pairs $\langle s,t\rangle$, and evaluation set with source sentences $\mathbf{s}$ and reference translations $\mathbf{r}$, with annotations on the paired terms (the ones appearing in the terminology) over both sides. 

In particular, we will solely rely on the annotations over the test sets, which we will assume are the comprehensive set of terms that need to be taken into account for evaluating the MT output. We do so to avoid complications arising from cases of untranslated source terms or target terms with no corresponding source-side terms. Such cases are not uncommon; for instance, a source-side term could be translated without adhering to the terminology target translation due to different context or domain. An example of such annotation is shown in \Tref{tab:example}; in practice the terms are marked with corresponding XML tags.

To create such annotations over the TICO-19 benchmark, we first find terminology matches over the parallel data, which are then verified by professional translators. In a more general case, one could first find terminology matches over an existing dataset, mark them, and treat them as gold-standard\footnote{Or silver-standard, if the process is assumed to be somewhat noisy.} annotation.

\subsection{Metrics}
Before discussing our proposed metrics, we outline the desirable properties of such a terminology-focused metric. The metric should reveal: 
\begin{enumerate}
    \item if a term has been translated according to specifications (\textit{adequacy}),
    \item if the term is placed in the appropriate context  (roughly corresponding to \textit{fluency})
\end{enumerate}

We clarify that we intend for such a metric to \emph{complement} other segment-level or corpus-level scores that evaluate an MT system. A holistic view that combines the general and the term-focused metrics would allow us to distinguish between (a) good translation systems that also adhere to the terminology specifications, (b) good translation systems that do not follow the terminology, (c) worse translation systems that nevertheless do adhere to the specifications, or (d) generally bad translation systems that also do not properly handle terms.

We first discuss several options for evaluating a system's output focusing on terminologies, also identifying potential advantages and disadvantages.

\paragraph{Exact-Match Accuracy} 
Perhaps the simplest option is to use a standard aggregate accuracy over each term, and if its terminology-defined translation appears in the output, we count it as correct. 

\[
\text{accuracy}(\mathbf{h},\mathbf{r}, \mathcal{T}) = \frac{\#\text{matched source terms}}{\#\text{source terms}}.
\]

In the case where a source term appears two or more times in the source sentence, they cannot be matched with the same hypothesis term. 
In the example of Table~\ref{tab:example}, the matching terms are fever--fiebre, cough--tos, dry cough--tos seca, and symptoms--s\'intomas. The first MT system produces all the target term translations, so it would get a $\frac{4}{4} = 100\%$ accuracy, while the second would get $\frac{3}{4} = 75\%$.
This metric is easy to compute, but it can be ``cheated" by e.g. simply appending all source term translations at the beginning or end of the system's hypothesis. 

\paragraph{Window Overlap} The exact-match accuracy by itself can only give us an indication of  whether the term appears in the output according to the desired specifications. It cannot, however, evaluate whether this target term is correctly placed in the output sentence. To address this shortcoming, we propose a simple window overlap metric that will complement the exact-match accuracy.

In particular, for each exact-matched target term (which, thus, will appear in both the hypothesis and the reference) we will define a window of $n$ content words (ignoring punctuation and stopwords) to the left and right of the term. We will then compute the percentage of tokens from this window that intersect between the hypothesis and the reference window, without counting the term words in the calculation.

\paragraph{Terminology-biased TER} Translation Edit Rate \cite[\textsc{ter}]{snover2006study} is an edit-distance based metric, which goes beyond insertions/deletions/substitutions by incorporating shifts that minimimze the total net cost, which intuitively account for possible valid reorderings. We propose a straightforward modification to TER, named \textsc{TERm}, where errors that concern the terminology tokens are penalized more than other tokens. Simply put, any reference word that belongs to a target term will have an edit cost of  $C_{\text{term}}>1$ instead of the standard cost of one. In our experiments, we set $C_{\text{term}}$ to 2.

To enhance comparability with other metrics and for simplicity reasons we report $1-\mathrm{TERm}$, so that a higher score will be better.

\section{Experiments and Results}

\paragraph{Test Suite} The TICO-19 test suite~\cite{anastasopoulos-etal-2020-tico} was developed to evaluate how well can MT systems handle the newly emerged topic of COVID-19. Accurate automatic translation can play an important role in facilitating communication in order to protect at-risk populations and combat the infodemic of misinformation, as described by the World Health Organization.
The dataset provides manually created translations of COVID-19 related data. The test set consists of PubMed articles (678 sentences from 5 scientific articles), patient-medical professional conversations (104 sentences), as well as related Wikipedia articles (411 sentences), announcements (98 sentences from Wikisource), and news items (67 sentences from Wikinews), for a total of \numprint{2100} sentences.

\begin{table*}[t]
    \centering
    \begin{tabular}{lp{3cm}|p{2cm}p{2cm}p{3.5cm}p{2cm}}
    \toprule
        &System & BLEU $\uparrow$ & Term: Exact Match $\uparrow$ & Term: Window Overlap (Window 2/3) $\uparrow$ & $1-\mathrm{TERm}$ $\uparrow$ \\
    \midrule
    \multicolumn{5}{l}{\textbf{En-Fr}} \\
        &\textsc{opus} & 46.24 & 82.22 & 60.12/58.90 & 61.08  \\
        &\textsc{fairseq} & 46.11 & 81.64  & 60.05/58.44 & 59.77  \\
        &\textit{\textsc{opus}-cheating} & \textit{46.40} & \textit{100.0} & \textit{57.17/56.63} & \textit{60.71} \\
        &\textit{\textsc{fairseq}-cheating} & \textit{45.85} & \textit{100.0} & \textit{56.82/56.04} & \textit{59.34} \\
    \midrule
    \multicolumn{5}{l}{\textbf{En-Ru}} \\
        &\textsc{opus} & 25.47 & 71.55 & 38.79/38.68 & 40.47  \\
        &\textsc{fairseq} & 28.88 & 77.29 & 42.27/42.47 & 44.69  \\
        &\textit{\textsc{opus}-cheating} & \textit{25.62} & \textit{100.0} & \textit{36.42/37.26} & \textit{40.02} \\
        &\textit{\textsc{fairseq}-cheating} & \textit{29.01} & \textit{100.0} & \textit{40.51/41.62} & \textit{44.26} \\
    \midrule
    \multicolumn{5}{l}{\textbf{Fr-En}} \\
        &\textsc{opus} & 39.43 & 74.84 & 35.75/36.16 & 48.60 \\
        &\textit{\textsc{opus}-cheating} & \textit{39.04} & \textit{100.0} & \textit{34.70/35.52} & \textit{47.92} \\
        &\textsc{naver}-general & 38.96 & 61.51 & 32.41/32.37 & 49.31 \\
        &\textsc{naver}-medical & 39.42 & 62.15 & 32.97/32.91 & 49.71 \\
    \midrule
    \multicolumn{5}{l}{\textbf{Ru-En}} \\
        &\textsc{opus} & 29.02 & 85.51 & 33.50/33.77 & 42.03 \\
        &\textit{\textsc{opus}-cheating} & \textit{28.91} & \textit{100.0} & \textit{32.09/32.63} & \textit{41.45} \\
        &\textsc{fairseq}-unconstr. & 32.85 & 73.58 & 46.14/46.54 & 43.93 \\
        &\textit{\textsc{fairseq}-constr.} & 28.26 & 73.71 & 40.33/40.93 & 30.02 \\
    \bottomrule
    \end{tabular}
    \caption{Results on TICO-19 dataset with \textsc{opus}, \textsc{fairseq} \& \textsc{naver} systems. Going beyond BLEU with the window overlap and $1-\mathrm{TERm}$ metrics is required because they are the only metrics that penalize cheating systems (results with cheating systems are \textit{italicized}).}
    \label{tab:my_label1}
\end{table*}

\paragraph{TICO-19 Systems and Results}
We first evaluate the systems used by \citet{anastasopoulos-etal-2020-tico} to benchmark the TICO-19 dataset. These include:
\begin{itemize}
    \item the OPUS-MT systems~\cite{TiedemannThottingal:EAMT2020} which are trained on the OPUS parallel data~\cite{tiedemann2012parallel} using the Marian toolkit~\cite{junczys2018marian};
    \item For English--Russian we compare against the pre-trained Fairseq models that won the WMT Shared Task in this direction last year~\cite{ng-etal-2019-facebook}, as well as the English to French system of \citet{ott-etal-2018-scaling}.
    \item We also use the many-to-one multilingual NMT model released by NAVER \cite{berard-etal-2020-multilingual} which translates from French, German, Italian, Spanish, and Korean into English. It was trained on a large amount parallel data from various domains, and produces  SOTA- or near-SOTA-level results for news, IWSLT, and biomedical translation tasks.
\end{itemize}
We present results with these systems in Table~\ref{tab:my_label1}. In addition to plain translations, we report the evaluation results for ``cheated" translations where the not-produced target terms were appended at the end of the output for each system. 

We first note that BLEU scores do not significantly penalize the "cheating approach". For example, the cheating En-Fr \textsc{opus} system achieves a slightly higher BLEU score over its non-cheating counterpart (c.f. 46.40 and 46.24). As expected the cheating systems easily achieve a 100\% exact-match accuracy, but they are penalized by the window overlap metric. This means that, although the cheating systems produce all desired term outputs, they do not correctly place them in the output. As such, we can be confident that a combination of BLEU, exact-match accuracy, and window overlap, can provide enough information to adequately evaluate the systems' handling of terminologies.

In addition, we find that the $1 - \mathrm{TERm}$ metric can also provide similar information in a complementary fashion. The cheating systems are indeed penalized by $\mathrm{TERm}$ (as the appended terms need to be shifted to be correctly matched, and these shifts have an edit cost of $C_{\text{term}}$).

\paragraph{Evaluation of WMT'20 systems}
The systems that were submitted to WMT 2020 were evaluated using
the test set from the TICO-19 dataset. Corpus-level evaluation on TICO-19 for multiple language pairs (using BLEU) was presented in~\citet{barrault-EtAl:2020:WMT1}. Here we expand by additionally evaluating these systems using our terminology-targeted metrics.

The English-Russian results are outlined in Table~\ref{tab:wmt-enru}, while the English-Chinese ones are shown in Table~\ref{tab:wmt-enzh}. We also provide English-Khmer (Table~\ref{tab:wmt-enkm}), English-Pashto (Table~\ref{tab:wmt-enps}), and English-Tamil (Table~\ref{tab:wmt-enta}) results.

In English-Russian, \texttt{online-b} is the best system, both in terms of overall BLEU score as well as over our term-focused metrics. The necessity for going beyond just exact-match accuracy for terminologies is showcased by the comparison between the \texttt{online-b} and \texttt{PROMT\_NMT} systems, which achieve the same term exact-match accuracy (around 88\%) but the former is better when we also take window overlap into account. The two systems' scores for $\mathrm{TERm}$ also confirm that intuition. 

We find a very interesting case in the English-Khmer results, where the \texttt{Huoshan\_Translate} system is significantly better than all other systems in handling terminologies, with a term exact-match accuracy of 66\% (and window overlap of 26\%) with the second best system \texttt{online-b} reaching almost half the scores (39\% accuracy, 19\% window overlap). On the other hand, the BLEU score for \texttt{online-b} is almost 15 points higher than \texttt{Huoshan\_Translate}! This means that \texttt{online-b} is generally a better MT system, but it does not adhere to the terminology\footnote{We note that this is a post-hoc analysis. The terminologies were not provided in advance to the shared task participants, so there is no a priori expectation or requirement that the systems adhere to the terminology.} while \texttt{Huoshan\_Translate} is a generally worse system which, however, is consistent in translating the terminology terms.

\begin{figure*}[t]
    \centering
    \begin{tabular}{cc}
        \includegraphics[width=.45\textwidth]{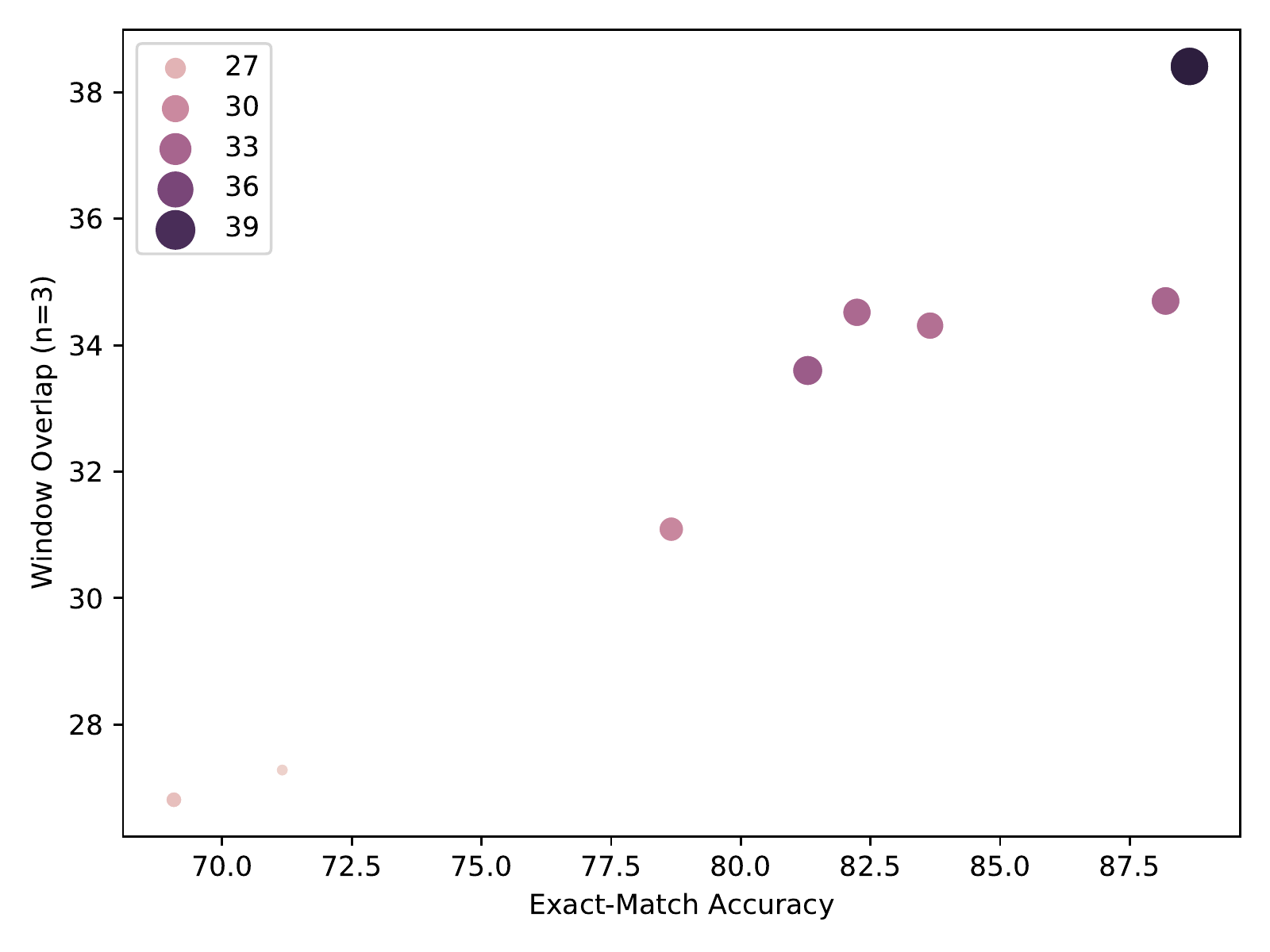} & \includegraphics[width=.45\textwidth]{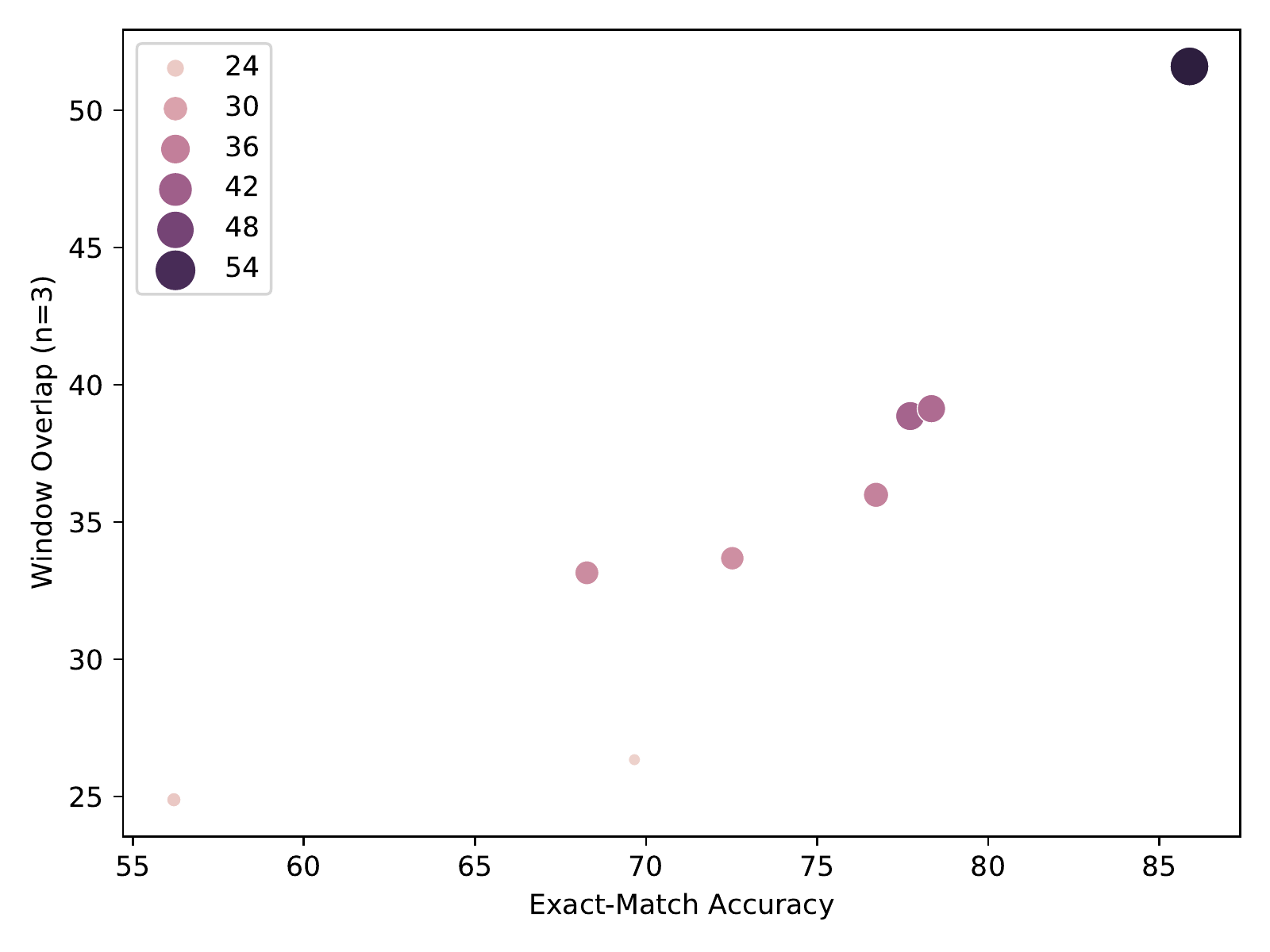}  \\
        WMT'20 English-Russian systems & WMT'20 English-Chinese systems
    \end{tabular}
    \caption{Term exact-match accuracy correlates with term Window overlap (examples from two language pairs).}
    \label{fig:viz}
\end{figure*}

For the other language pairs the clear winning system of the shared task is also the one that better handles terminology terms (the \texttt{SJTU-NICT} system for English-Chinese, and  \texttt{online-b} for English-Pashto and English-Tamil). 
We note that while one would expect that the best overall translation would also be the best in terms of terminology translation, there is only a mild correlation between BLEU scores and the term exact-match accuracy scores. 
We compute the Spearman's rank correlation between BLEU score and two measures, window-Overlap of size 3 and exact match (see Table~\ref{tab:correlation}).
On the other hand, we do observe a high correlation between the term exact-match accuracy and the window overlap scores. Figure~\ref{fig:viz} presents these correlations for the EN-RU and EN-ZH WMT systems. These implies that, encouragingly, a MT system that is good at handling terminology is also generally good at correctly placing the terms in the appropriate context.

\begin{table*}[t]
    \centering
    \begin{tabular}{c|p{2cm}p{2cm}p{3.5cm}p{2cm}}
    \toprule
        System & BLEU (standard) $\uparrow$ & Term: Exact Match $\uparrow$  & Term: Window Overlap (Window 2/3) $\uparrow$ &  $1-\mathrm{TERm}$ $\uparrow$\\
    \midrule
        online-b & 40.74 & 88.65 & 38.93/38.41 & 55.92 \\
        Online-G & 33.84 & 81.29 & 33.78/33.60  & 50.73\\
        PROMT\_NMT & 32.92 & 88.19 & 34.56/34.70 & 50.40 \\
        ariel197197 & 32.68 & 82.24 & 34.64/34.52 & 49.88 \\
        OPPO & 32.05 & 83.65 & 34.28/34.31  & 50.31\\
        online-a & 30.23 & 78.66 & 31.22/31.09 & 46.85 \\
        zlabs-nlp & 26.14 & 69.07 & 26.88/26.81 & 43.48 \\
        online-z & 24.88 & 71.16 & 27.17/27.28 & 44.28 \\
    \bottomrule
    \end{tabular}
    \caption{WMT'20 system results for English-Russian.}
    \label{tab:wmt-enru}
\end{table*}
\begin{table*}[t]
    \centering
    \begin{tabular}{c|p{2cm}p{2cm}p{3.5cm}p{2cm}}
    \toprule
        System & BLEU (standard) $\uparrow$ & Term: Exact Match $\uparrow$ & Term: Window Overlap (Window 2/3) $\uparrow$ & $1-\mathrm{TERm}$  $\uparrow$ \\
    \midrule
        SJTU-NICT & 57.82 & 85.88 & 51.92/51.60 & 70.30 \\
        OPPO & 40.82 & 77.72 & 38.94/38.86 & 52.55 \\
        online-b & 39.55 & 78.34 & 38.90/39.13 & 54.98 \\
        online-a & 35.23 & 76.72 & 35.69/35.99 & 50.76 \\
        WMT & 33.72 & 68.27 & 32.93/33.15 & 47.16 \\
        Online-G & 33.17 & 72.52 & 33.67/33.68 & 47.80  \\
        zlabs-nlp & 24.17 & 56.20 & 24.52/24.88 & 38.05 \\
        online-z & 22.69 & 69.66 & 26.03/26.34 & 36.16 \\
    \bottomrule
    \end{tabular}
    \caption{WMT'20 system results for English-Chinese.}
    \label{tab:wmt-enzh}
\end{table*}
\begin{table*}[t]
    \centering
    \begin{tabular}{c|p{2cm}p{2cm}p{3.5cm}p{2cm}}
    \toprule
        System & BLEU (standard) $\uparrow$ & Term: Exact Match $\uparrow$ & Term: Window Overlap (Window 2/3) $\uparrow$ & $1-\mathrm{TERm}$  $\uparrow$ \\
    \midrule
        online-b & 35.54 & 38.84 & 19.20/18.07 & 53.82 \\
        online-z & 20.79 & 23.91 & 08.95/08.17 & 38.18 \\
        Huoshan\_Translate & 19.58 & 66.05 & 26.16/23.84 & 36.84 \\
        OPPO & 19.85 & 14.00 & 06.53/05.77 & 36.99 \\
        Online-G & 14.33 & 01.59 & 00.74/00.72 & 34.04 \\
    \bottomrule
    \end{tabular}
    \caption{WMT'20 system results for English-Khmer.}
    \label{tab:wmt-enkm}
\end{table*}
\begin{table*}[t]
    \centering
    \begin{tabular}{c|p{2cm}p{2cm}p{3.5cm}p{2cm}}
    \toprule
        System & BLEU (standard) $\uparrow$ & Term: Exact Match $\uparrow$ & Term: Window Overlap (Window 2/3) $\uparrow$ & $1-\mathrm{TERm}$  $\uparrow$ \\
    \midrule
        online-b & 39.88 & 84.96 & 46.32/45.59 & 51.97\\
        OPPO & 21.47 & 59.92 & 24.63/23.44 & 36.97\\
        Huoshan\_Translate & 20.88 & 70.80 & 30.65/29.39 & 39.68 \\
        online-z & 16.42 & 51.96 & 19.07/18.43 & 35.92\\
    \bottomrule
    \end{tabular}
    \caption{WMT'20 system results for English-Pashto.}
    \label{tab:wmt-enps}
\end{table*}
\begin{table*}[t]
    \centering
    \begin{tabular}{c|p{2cm}p{2cm}p{3.5cm}p{2cm}}
    \toprule
        System & BLEU (standard) $\uparrow$ & Term: Exact Match $\uparrow$ & Term: Window Overlap (Window 2/3) $\uparrow$ & $1-\mathrm{TERm}$  $\uparrow$ \\
    \midrule
        online-b & 30.30 & 76.48 & 29.56/29.78 & 50.28 \\
        Facebook\_AI & 15.57 & 41.89 & 13.56/13.48 & 34.06 \\
        online-a & 14.48 & 49.66 & 15.12/14.68 & 33.22 \\
        OPPO & 12.82 & 31.96 & 10.16/10.07 & 31.89 \\
        UEDIN & 12.21 & 44.63 & 11.99/11.76 & 28.24 \\
        Microsoft\_STC\_India & 11.93 & 30.37 & 08.53/08.70 & 28.96 \\
        online-z & 11.62 & 41.89 & 10.93/10.74 & 31.12 \\
        Huoshan\_Translate & 11.61 & 39.38 & 11.68/11.52 & 29.62 \\
        zlabs-nlp & 10.32 & 40.87 & 09.72/09.49 & 23.26 \\
        DCU & 9.74 & 29.00 & 07.03/06.98 & 18.79 \\
        Groningen & 8.94 & 51.94 & 10.46/10.76 & 24.22 \\
        Online-G & 7.32 & 48.52 & 09.43/09.57 & 18.39 \\
        TALP\_UPC & 6.30 & 24.32 & 04.70/04.58 & 14.51 \\
        SJTU-NICT & 2.93 & 24.89 & 03.22/03.42 & 15.97 \\
    \bottomrule
    \end{tabular}
    \caption{WMT'20 system results for English-Tamil.}
    \label{tab:wmt-enta}
\end{table*}
\begin{table*}[t]
    \centering
    \begin{tabular}{lp{3cm}|p{2cm}p{2cm}p{3.5cm}p{2cm}}
    \toprule
        &System & BLEU $\uparrow$ & Term: Exact Match $\uparrow$ & Term: Window Overlap (Window 2/3) $\uparrow$ & $1-\mathrm{TERm}$ $\uparrow$ \\
    \midrule
    \multicolumn{5}{l}{\textbf{En-De}} \\
        &\textsc{baseline} & 22.41 & 33.83 & 19.67/19.13 & 22.90 \\
        &\textsc{tla} & 29.17 & 89.22 & 45.86/45.93 & 41.96 \\
    \midrule
    \multicolumn{5}{l}{\textbf{En-Et}} \\
        &\textsc{baseline} & 15.49 & 45.73 & 19.93/20.23 & 29.29 \\
        &\textsc{tla} & 17.02 & 76.22 & 37.50/36.19 & 32.39 \\
    \midrule
    \multicolumn{5}{l}{\textbf{En-Lt}} \\
        &\textsc{baseline} & 23.28 & 56.76 & 32.28/32.60 & 35.10 \\
        &\textsc{tla} & 27.91 & 88.06 & 51.51/51.57 & 46.21 \\
    \midrule
    \multicolumn{5}{l}{\textbf{En-Lv}} \\
        &\textsc{baseline} & 28.82 & 59.32 & 37.08/37.52 & 48.38 \\
        &\textsc{tla} & 27.14 & 85.03 & 44.70/45.84 & 49.26 \\
    \bottomrule
    \end{tabular}
    \caption{Results on ATS dataset with \textsc{baseline} \& \textsc{tla} systems.}
    \label{tab:toms_marcis}
\end{table*}

\begin{table*}[t]
    \centering
    \begin{tabular}{c|cc|c}
    \toprule
        & \multicolumn{2}{c|}{Exact Match} & \\
        Language & (Original Dataset)  & (Lemmatized Dataset) & Total Matches \\
    \midrule
        French & 2945 & 649 & 3594 \\
        Khmer & 1021 & N/A & 1021\\
        Pashto & 982 & N/A & 982\\
        Russian & 1652 & 1395 & 3047\\
        Tamil & 987 & 182 & 1169\\
        Chinese & 2819 & N/A & 2819 \\
    \bottomrule
    \end{tabular}
    \caption{Exact terminology matches found on Tico dataset.}
    \label{tab:exactmatch}
\end{table*}

\begin{table*}[t]
    \centering
    \begin{tabular}{l|cc}
\toprule
lang & Term: Win-3 & Term: Exact Match \\
\midrule
rus & $0.8333^*$ & $0.8095^*$ \\
zho & $0.9286^*$ & $0.8571^*$ \\
tam & $0.6271^*$ & $0.4154$ \\
khm & -- & -- \\
pus & -- & -- \\
\bottomrule
\end{tabular}
    \caption{Spearman Rank correlation for \texttt{ru} (Table~\ref{tab:wmt-enru}), \texttt{zh} (Table~\ref{tab:wmt-enzh}), \texttt{ta} (Table~\ref{tab:wmt-enta}), \texttt{khm} (Table~\ref{tab:wmt-enkm}) and \texttt{pash} (Table~\ref{tab:wmt-enps}). If the $p$-value is higher than 0.2 we omit the correlation value. A * indicates a $p$-value lower than 0.1. }
    \label{tab:correlation}
\end{table*}

\begin{figure*}
    \centering
    \begin{tabular}{ccc}
    \toprule
    \multicolumn{3}{c}{English-French: Is the term translated correctly? (support: 90)}\\
        \includegraphics[width=.3\textwidth]{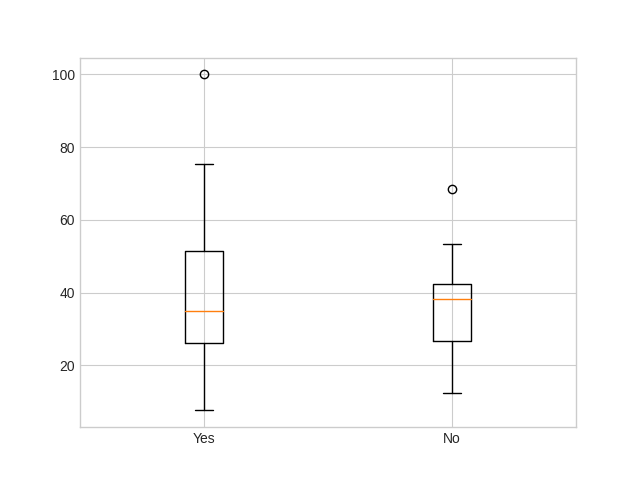} & \includegraphics[width=.3\textwidth]{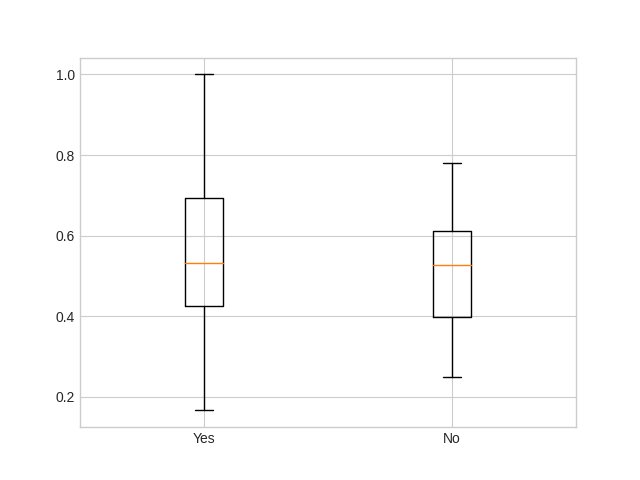} &
        \includegraphics[width=.3\textwidth]{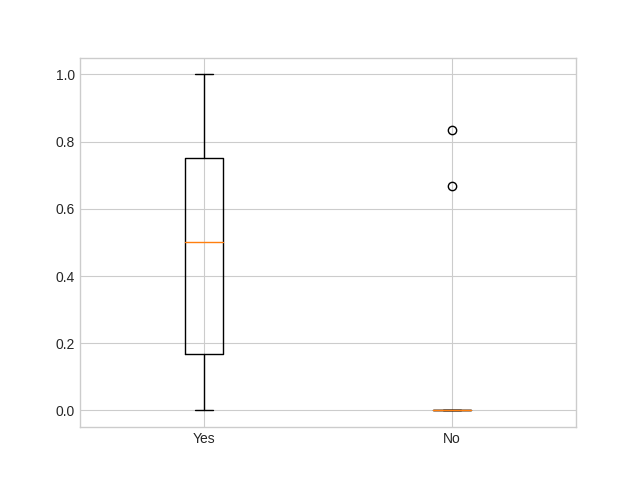} \\
        \small BLEU & \small 1 - TERm & \small Window overlap \\
    \midrule
            \multicolumn{3}{c}{English-Russian: Is the term translated correctly? (support: 190)}\\
            \includegraphics[width=.3\textwidth]{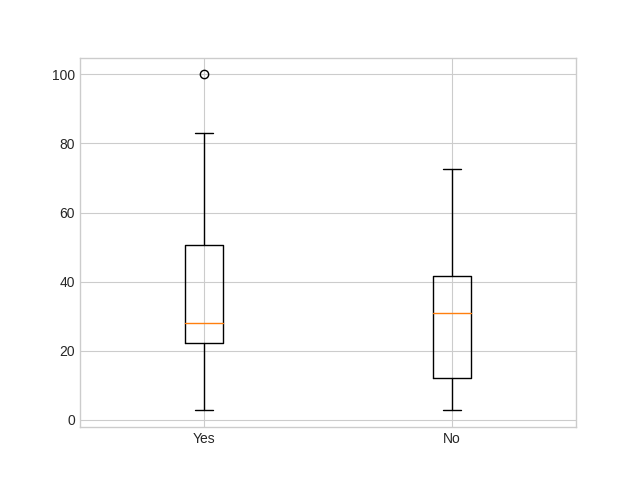} & \includegraphics[width=.3\textwidth]{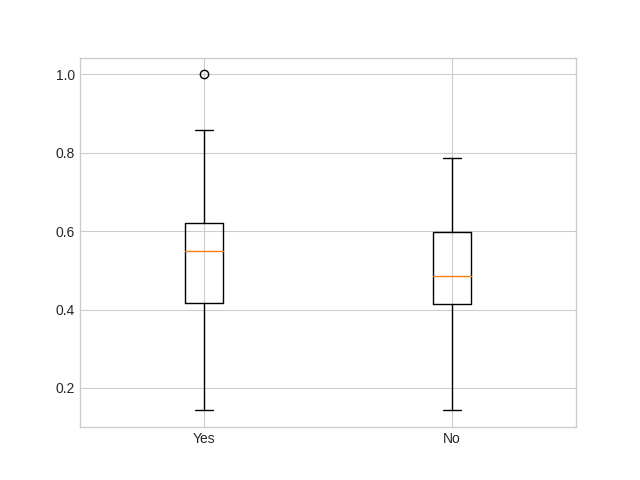} &
            \includegraphics[width=.3\textwidth]{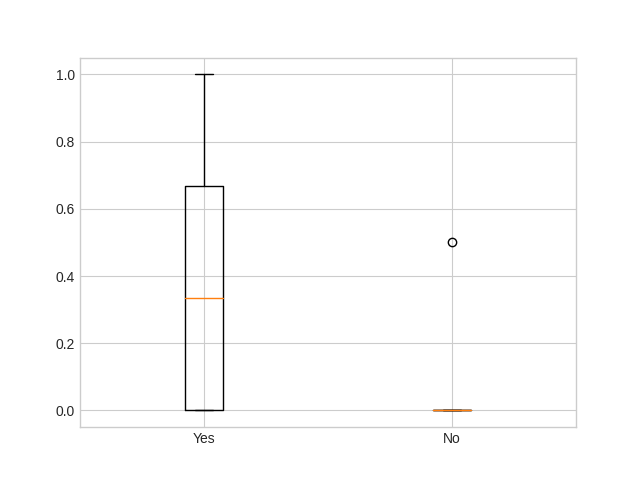} \\
        \small BLEU & \small 1 - TERm & \small Window overlap \\
    \bottomrule
    \end{tabular}
    \caption{Aggregate scores for outputs annotated as having correctly translated terms (``Yes") and incorrectly translated terms (``No"). The combination of Window Overlap with other sentence-level metrics can distinguish good from bad term translations.}
    \label{fig:box}
\end{figure*}

\begin{figure*}
    \centering
    \begin{tabular}{cc}
    \toprule
        \multicolumn{2}{c}{English-French} \\
        \includegraphics[width=.33\textwidth]{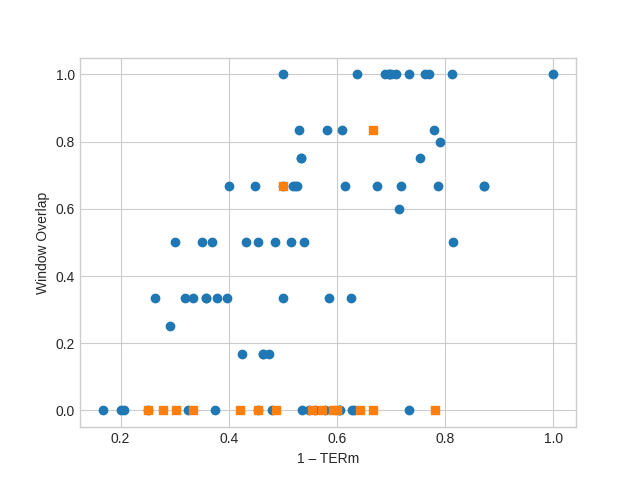} &
        \includegraphics[width=.33\textwidth]{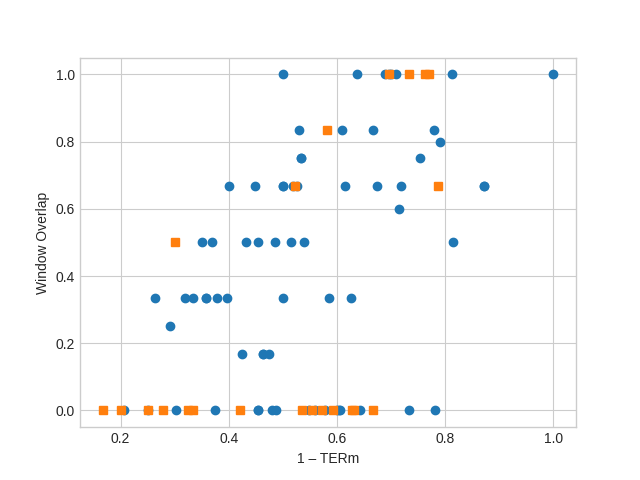} \\
    \small Is the term translation correct? & \small Is the term's context correct? \\
    \midrule
    \multicolumn{2}{c}{English-Russian} \\
        \includegraphics[width=.33\textwidth]{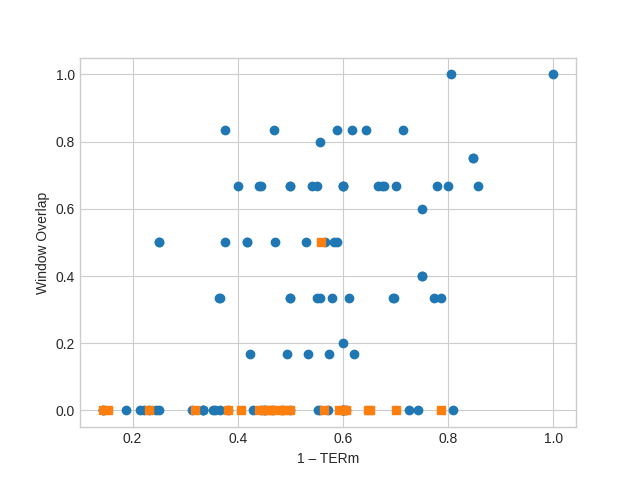} &
        \includegraphics[width=.33\textwidth]{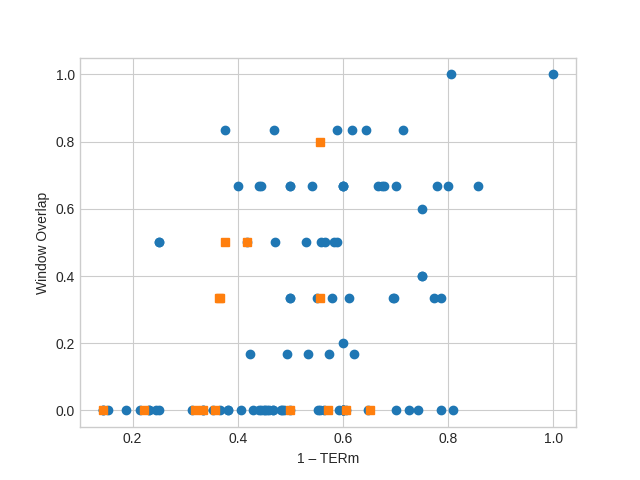} \\
    \small Is the term translation correct? & \small Is the term's context correct? \\
    \bottomrule
    \end{tabular}
    \caption{Scatter-plot of 1 - TERm vs Window Overlap for our human annotator surveys. Blue points correspond to the question's answer being ``yes" and orange points received a ``no" answer. Despite a few false positives and false negatives, the combination of 1 - TERm and window overlap can capture the quality of terminology translation.}
    \label{fig:scatter}
\end{figure*}

\paragraph{Ternminology-Augmented Systems} Last, we test the application of our metrics on the MT outputs of the recent work of~\citet{bergmanis2021facilitating}, which presents a data augmentation method for training MT systems that can handle terminology translations using only target lemma annotations (as opposed to being provided with the correct inflected word in advance). The results for the baseline system and the proposed TLA system in four language pairs (English to German, Estonian, Lithuanian, and Latvian) are listed in Table~\ref{tab:toms_marcis}.\footnote{We thank the authors for providing the outputs of their systems.} We find that in all cases the TLA systems are significantly better in handling terminologies, both in terms of exact match accuracy as well as term window overlap and TERm. As before, such changes are not immediately noticeable using only corpus- or sentence-level metrics like BLEU (cf results in En-Lv).

\section{Human Annotator Study}
We conducted a small-scale human annotation study on English-French and English-Russian MT outputs. We presented experienced bilingual speakers with the English source sentence, the source term and its expected target translation, along with the MT output. The annotators were tasked with answering two yes/no questions:
\begin{enumerate}
    \item Is the source term translated correctly? 
    \item Is the term's surrounding context correct in the translation?
\end{enumerate}

We used the outputs from the two best English-French systems (two top in Table~\ref{tab:my_label1})  and the three best English-Russian systems (first three in Table~\ref{tab:wmt-enru}). In total, we annotated 90 outputs in French and 192 outputs in Russian. Despite the small sample sizes, we nevertheless find that a combination of our metrics can effectively distinguish correctly translated terms from term translations that are wrong (or missing). Figure~\ref{fig:box} presents box-plots of three metrics (BLEU, TERm, and term window overlap) for the two subsets of MT outputs as produced by our annotators when answering the first question (``Is the source term translated correctly?"). The visualizations show that BLEU or TERm by themselves are inadequate for this task as there's no observable difference between the two classes. Including term window overlap, however, we find significant differences between the classes for both languages.

This observation is more clear in the scatter plots in Figure~\ref{fig:scatter}, which show the combination of TERm and window overlap, color-coded according to the annotators' answers to our two survey questions (blue for ``yes", orange for ``no"). We find that the vast majority of incorrect or missing term translations receive very low window overlap scores, with only a few false positives (i.e. wrong translations that receive high scores) or false negatives (i.e. correct translations receiving low scores).

\section{Conclusion}
We present methods for evaluating the consistency of MT outputs on term translation, given a terminology to which the MT systems must adhere. We showcase and analyze the proposed metrics on three different datasets, concluding that a combination of sentence-level scores (TERm) and terminology-targeted evaluation (exact match accuracy plus term window overlap) can reveal the MT systems that are better, or more consistent, in terms of terminology translation. We also confirm our findings through a limited human annotation study. Code for computing all metrics discussed in the paper is publicly available.\footnote{\url{https://github.com/mahfuzibnalam/terminology_evaluation}}

\section*{Acknowledgements} The authors are thankful to Georgiana Dinu and Marcello Federico for helpful discussions on the metrics formulations, as well as for feedback on preliminary versions of this work.

\bibliographystyle{acl_natbib}
\bibliography{acl2021}

\clearpage
\newpage
\appendix
\section{Other Potential Metrics}
\paragraph{Partial-Match Accuracy} 
An extension of the exact match accuracy, this will take into account the number of words in each target term and allow for partial matches. For each source-target term pair $\langle s,t\rangle$ as they appear in a source-reference sentence pair $(\mathbf{s},\mathbf{r})$, we define
\begin{align*}
    \text{match}(t,\mathbf{h},\mathbf{r}) &= \frac{\sum_{w \in t} \delta(w\in \mathbf{h})}{|t|},\\
    \text{accuracy}(\mathbf{h},\mathbf{r}, \mathcal{T}) &= \frac{\sum_{\langle s,t \rangle \in (\mathbf{s},\mathbf{r})} \text{match}(t,\mathbf{h},\mathbf{r})}{\#\text{source terms}}.
\end{align*}

Continuing on the example from Table~\ref{tab:example}, the source term ``dry cough" needs to be matched to two words in Spanish (``tos seca"). MT system 1 indeed matches both words of that term, so the its partial overall accuracy doesn't change ($\frac{4}{4} = 100\%$). The second MT system, though, has a partial match for the ``dry cough--tos seca" term, matching the word ``tos". As a result, its overall partial-match accuracy will be $\frac{3.5}{4} = 87.5\%$.

This metric is similarly easy to compute as exact-match accuracy, and it can be ``cheated" in a similar manner. In addition, the partial-match accuracy could potentially overestimate how well a system performs, by rewarding it for common words that are part of multi-word terms. For instance, in the first term shown in Table~\ref{tab:terminology}, two of the four words of the Spanish translation are highly frequent ones (``la" and ``de") which could appear elsewhere in the translation. A potential solution could follow the paradigm of metrics like METEOR~\cite{banerjee2005meteor} by upweighting content words and down-weighting stopwords; such information is not widely available for all languages. 

\paragraph{Alignment-based partial-match accuracy} The previous metrics simply search for term translation matches throughout the hypothesis, which can lead to misleading results. 
A solution would be to instead rely on specific alignments.

We define an alignment $\mathcal{A}$ as a set of word-level pairs $\{\langle s_i, h_j \rangle \}$ source words $s_i \in \mathbf{s}$ and hypothesis words $h_i \in \mathbf{h}$.
The alignment in its general case allows for many-to-many, as well as one-to-many and one-to-one alignments.

We can now define a $\alpha(t, \mathbf{s,h}, \mathcal{A})$ function which, given a (possibly multi-word) source term and the alignment between the two sentences, returns the set of hypothesis words that are aligned to the words of the term. 
We now modify our matching function such that we only search for matches constrained by the alignment
\begin{align*}
    \text{match}(t,\mathbf{h},\mathbf{r}) &= \frac{\sum_{w \in \alpha(t)} \delta(w\in \mathbf{h})}{|t|}
\end{align*}
and the accuracy computation remains the same.

This approach will \textit{probably} deal with the game-ability of the previous two metrics, as long as the alignment is biased to not allow too much reordering (e.g. a prior like the diagonal prior used in fast\_align might help), but this could also be an issue for some language pairs. The main issue with this approach is going to be the potential mistakes made by the alignment mechanism. 

\paragraph{UD-alignment-based partial-match accuracy} Exactly as before, with the exception that the window of the term is not defined in absolute terms (e.g. previous and following two words) but is instead defined as all words connected to words of the term through dependency edges, as obtained by a dependency parser. We use Stanza pre-trained dependency parsing models for our experiments~\cite{qi2020stanza}.

\paragraph{Reordering penalty for partial match} Another direction that would limit the game-ability of the exact- or partial-match accuracy approaches is to add a reordering penalty. Both of the first two metrics are not affected at all by the positioning of the terms within the hypothesis sentence. 

Ideally, we would want the MT system to produce the term's translation in approximately the same relative output sentence position, so we can add a multiplicative penalty as follows.
The relative position of a word $w_i \in \mathbf{s}$ is simple going to be $\frac{i}{|s|}$; as such the relative position of any word will be a real number between 0 (the first word) and 1 (the last word in the sentence). For a word $w$ that appears both in the hypothesis and the reference in positions $i$ and $j$ respectively, we now define the reordering penalty
\begin{align*}
\text{RP}(w,\mathbf{h},\mathbf{r} :\ &w_i\in\mathbf{h}, w_j\in\mathbf{r}, w=w_i=w_j) = \\
&= 1 - \left| \frac{i}{|\mathbf{h}|} - \frac{j}{|\mathbf{r}|} \right|. 
\end{align*}
If the word appears in very similar relative positions in the two sentences, the reordering penalty will be minimal (close to 1).
We now simply modify our matching function to include the reordering penalty:
\begin{align*}
    \text{match}(t,\mathbf{h},\mathbf{r}) = \frac{\sum_{w \in t} \delta(w\in \mathbf{h})\cdot\text{RP}(w,\mathbf{h},\mathbf{r})}{|t|},\\
    \text{accuracy}(\mathbf{h},\mathbf{r}, \mathcal{T}) = \frac{\sum_{\langle s,t \rangle \in (\mathbf{s},\mathbf{r})} \text{match}(t,\mathbf{h},\mathbf{r})}{\#\text{source terms}}.
\end{align*}

This metric is as easily computed as the exact- or partial-match accuracy, but the reordering penalty should take care of the potential to game the metrics by simply concatenating all target terms in the beginning or end of a hypothesis. However, one could fathom an adversarial player who instead inserts the terms throughout the hypothesis (perhaps in similar relative positions as the source terms, taking advantage of the fact that ``translationese" outputs tend to exhibit low rates of reordering), which might achieve lower reordering penalties in the average case than the naive concatenation. 

\paragraph{Term-BLEU} None of the above metrics take into account whether the target term, if it appears, is used within the appropriate context.
The metric we propose here will transform a hypothesis-reference pair into a set of phrases, centered around the terms using a left-right window of $k$ words, and then use common MT metrics (e.g. BLEU, BERTScore, COMET) over this new set of phrases.

We start by first aligning the hypothesis and the reference. Then, for a target term appearing in the reference in span $[r_i, r_j]$, we find its corresponding span $[h_m,\ldots,h_n]$ in the hypothesis using the word-level alignment. The corresponding span is defined as the span between the minimum and maximum positions in the reference sentence that have been aligned to the words of the  reference term (the alignments do not need to be contiguous). 
For each target term we construct a phrase pair $(r_{i-k}\ldots r_{j+k}, h_{m-k}\ldots h_{n+k}$, which will be treated as a secondary, term-centered, parallel corpus.

For example, for the Spanish reference in Table~\ref{tab:example} and the MT Output 2, we obtain an alignment between the two sentences, where the words fiebre and s\'intomas are exactly aligned, and ``tos seca" in the reference is both aligned to ``tos" in the output.
Since there are four terms in the gold data, the above process will create the following four parallel phrases, corresponding to each of the terms (using a window of $k=2$ as an example):
\begin{enumerate}
    \item[] Reference $|||$ Hypothesis 
    \item tambi\'en tenían \textbf{[fiebre]$_1$} (98\%) , $|||$ tambi\'en tenían \textbf{[fiebre]$_1$} (98\%) ,  
    \item (98\%) , \textbf{[tos]$_2$} seca (47\%) $|||$ (98\%) , tos (47\%) y 
    \item (98\%) , \textbf{[tos seca]$_3$} (47\%) y $|||$ (98\%) , tos (47\%) y 
    \item como principales \textbf{[síntomas]$_4$} . $|||$ sus principales s\'intomas .
\end{enumerate}
We can then use any MT metric over these parallel phrases, and produce a micro-average for all of them over the corpus. 

This metric has the advantage that it still rewards exact matches over the term and penalizes missing terms, but also takes into account the surrounding context.
It still has the disadvantage that it relies on alignment, but presumably monolingual alignments should be easier to get right than bilingual ones.
If we substitute BLEU with a different score (e.g. chrF or BERTScore) we should be able to also capture morphological variants.

\section{Results with Other Potential Metrics}
TICO results in Table~\ref{tab:app1} for the original systems, and in Tables~\ref{tab:app2} and~\ref{tab:app3} for the naive and smart cheating systems respectively.

\begin{table*}[t]
    \centering
    \begin{tabular}{p{2.25cm}|p{1cm}p{1cm}p{1.25cm}p{1cm}p{1.75cm}p{1.75cm}p{1.5cm}p{1.75cm}}
    \toprule
        Language Pair & BLEU (standard) $\uparrow$ & Term: Exact Match $\uparrow$ & Standard TER $\uparrow$ & Term-biased TER $\uparrow$ & Alignment-based: Exact Order Match (Window 2/ Window 3) $\uparrow$ & Alignment-based: BLEU (Window 2/ Window 3) $\uparrow$ & Alignment-based: UD Match $\uparrow$ & Term: Window Overlap (Window 2/ Window 3) $\uparrow$ \\
    \midrule
        en-fr (OPUS) & 46.24 & 82.22 & 61.37 & 61.08 & 59.50/55.08 & 53.84/49.03 & 59.08 & 60.12/58.90 \\
        en-fr (fairseq) & 46.11 & 81.64 & 60.08 & 59.77 & 59.24/54.74 & 53.74/48.77 & 59.19 & 60.05/58.44 \\
    \midrule
        en-ru (OPUS) & 25.47 & 71.55 & 40.96 & 40.47 & 38.58/34.92 & 33.79/29.91 & 34.42 & 38.79/38.68 \\
        en-ru (fairseq) & 28.88 & 77.29 & 45.08 & 44.69 & 41.97/38.37 & 36.77/33.73 & 37.83 & 42.27/42.47 \\
    \midrule
        fr-en (OPUS) & 39.43 & 74.84 & 49.14 & 48.60 & 33.79/31.63 & 30.32/27.73 & 35.77 & 35.75/36.16 \\
        ru-en (OPUS) & 29.02 & 85.51 & 42.46 & 42.03 & 30.18/27.08 & 27.69/23.65 & 33.68 & 33.50/33.77 \\
    \bottomrule
    \end{tabular}
    \caption{OPUS \& fairseq system results}
    \label{tab:app1}
\end{table*}
\begin{table*}[t]
    \centering
    \begin{tabular}{p{2.25cm}|p{1cm}p{1cm}p{1.25cm}p{1cm}p{1.75cm}p{1.75cm}p{1.5cm}p{1.75cm}}
    \toprule
        Language Pair & BLEU (standard) $\uparrow$ & Term: Exact Match $\uparrow$ & Standard TER $\uparrow$ & Term-biased TER $\uparrow$ & Alignment-based: Exact Order Match (Window 2/ Window 3) $\uparrow$ & Alignment-based: BLEU (Window 2/ Window 3) $\uparrow$ & Alignment-based: UD Match $\uparrow$ & Term: Window Overlap (Window 2/ Window 3) $\uparrow$ \\
    \midrule
        en-fr (OPUS) & 45.02 & 100.00 & 57.12 & 57.06 & 57.21/52.95 & 50.41/45.55 & 56.82 & 57.55/57.02 \\
        en-fr (fairseq) & 44.23 & 100.00 & 55.81 & 55.73 & 56.46/52.21 & 49.86/45.20 & 56.45 & 57.13/56.33 \\
    \midrule
        en-ru (OPUS) & 25.60 & 99.74 & 37.30 & 37.20 & 35.73/32.27 & 30.76/27.86 & 33.82 & 36.86/37.61 \\
        en-ru (fairseq) & 29.04 & 99.47 & 41.38 & 41.32 & 39.83/36.41 & 34.09/31.60 & 36.90 & 40.95/41.98 \\
    \midrule
        fr-en (OPUS) & 37.51 & 99.96 & 44.36 & 44.05 & 33.42/31.31 & 29.29/27.06 & 35.83 & 34.92/35.67 \\
        ru-en (OPUS) & 27.96 & 100.00 & 38.36 & 37.98 & 29.16/26.20 & 25.60/22.08 & 33.85 & 32.68/33.11 \\   
    \bottomrule
    \end{tabular}
    \caption{Naive cheating OPUS \& fairseq system results}
    \label{tab:app2}
\end{table*}
\begin{table*}[t]
    \centering
    \begin{tabular}{p{2.25cm}|p{1cm}p{1cm}p{1.25cm}p{1cm}p{1.75cm}p{1.75cm}p{1.5cm}p{1.75cm}}
    \toprule
        Language Pair & BLEU (standard) $\uparrow$ & Term: Exact Match $\uparrow$ & Standard TER $\uparrow$ & Term-biased TER $\uparrow$ & Alignment-based: Exact Order Match (Window 2/ Window 3) $\uparrow$ & Alignment-based: BLEU (Window 2/ Window 3) $\uparrow$ & Alignment-based: UD Match $\uparrow$ & Term: Window Overlap (Window 2/ Window 3) $\uparrow$ \\
    \midrule
        en-fr (OPUS) & 46.40 & 99.97 & 60.78 & 60.71 & 57.38/53.13 & 52.01/48.06 & 57.04 & 57.17/56.63 \\
        en-fr (fairseq) & 45.85 & 99.97 & 59.43 & 59.34 & 56.74/52.48 & 51.63/47.75 & 56.72 & 56.82/56.04 \\
    \midrule
        en-ru (OPUS) & 25.62 & 99.74 & 40.11 & 40.02 & 35.96/32.49 & 31.99/29.51 & 34.03 & 36.42/37.26 \\
        en-ru (fairseq) & 29.01 & 99.47 & 44.33 & 44.26 & 39.92/36.51 & 35.40/33.41 & 37.09 & 40.51/41.62 \\
    \midrule
        fr-en (OPUS) & 39.04 & 99.96 & 48.25 & 47.92 & 33.51/31.41 & 30.57/28.46 & 36.31 & 34.70/35.52 \\
        ru-en (OPUS) & 28.91 & 100.00 & 41.83 & 41.45 & 29.25/26.25 & 26.77/23.24 & 33.92 & 32.09/32.63 \\
    \bottomrule
    \end{tabular}
    \caption{Smart cheating OPUS \& fairseq system results}
    \label{tab:app3}
\end{table*}

\subsection{Constrained Decoding Systems}
We also evaluated the outputs of the constrained and unconstrained decoding using the Russian-English Fairseq system, with results presented in Table~\ref{tab:app4}.

\begin{table*}[t]
    \centering
    \begin{tabular}{p{2.25cm}|p{1cm}p{1cm}p{1.25cm}p{1cm}p{1.75cm}p{1.75cm}p{1.5cm}p{1.75cm}}
    \toprule
        Language Pair & BLEU (standard) $\uparrow$ & Term: Exact Match $\uparrow$ & Standard TER $\uparrow$ & Term-biased TER $\uparrow$ & Alignment-based: Exact Order Match (Window 2/ Window 3) $\uparrow$ & Alignment-based: BLEU (Window 2/ Window 3) $\uparrow$ & Alignment-based: UD Match $\uparrow$ & Term: Window Overlap (Window 2/ Window 3) $\uparrow$ \\
    \midrule
        ru-en (unconstrained) & 32.85 & 73.58 & 44.35 & 43.93 & 43.01/39.38 & 38.73/34.52 & 42.01 & 46.14/46.54 \\
        ru-en (constrained) & 28.26 & 73.71 & 30.47 & 30.02 & 35.90/32.63 & 32.37/28.96 & 35.18 & 40.33/40.93 \\
    \bottomrule
    \end{tabular}
    \caption{Unconstrained vs constrained fairseq system results}
    \label{tab:app4}
\end{table*}
\end{document}